# Measuring Praise and Criticism: Inference of Semantic Orientation from Association

PETER D. TURNEY
National Research Council Canada
and
MICHAEL L. LITTMAN
Rutgers University

The evaluative character of a word is called its *semantic orientation*. Positive semantic orientation indicates praise (e.g., "honest", "intrepid") and negative semantic orientation indicates criticism (e.g., "disturbing", "superfluous"). Semantic orientation varies in both direction (positive or negative) and degree (mild to strong). An automated system for measuring semantic orientation would have application in text classification, text filtering, tracking opinions in online discussions, analysis of survey responses, and automated chat systems (*chatbots*). This paper introduces a method for inferring the semantic orientation of a word from its statistical association with a set of positive and negative paradigm words. Two instances of this approach are evaluated, based on two different statistical measures of word association: pointwise mutual information (PMI) and latent semantic analysis (LSA). The method is experimentally tested with 3,596 words (including adjectives, adverbs, nouns, and verbs) that have been manually labeled *positive* (1,614 words) and *negative* (1,982 words). The method attains an accuracy of 82.8% on the full test set, but the accuracy rises above 95% when the algorithm is allowed to abstain from classifying mild words.

Authors' addresses: P.D. Turney, Institute for Information Technology, National Research Council Canada, M-50 Montreal Road, Ottawa, Ontario, Canada, K1A 0R6, email: peter.turney@nrc.ca; M.L. Littman, Department of Computer Science, Rutgers University, Piscataway, NJ 08854-8019, USA: email: mlittman@cs.rutgers.edu.

# 1. INTRODUCTION

In an early study of subjective meaning, Osgood *et al.* [1957] asked people to rate words on a wide variety of scales. Each scale was defined by a bipolar pair of adjectives, such as sweet/sour, rugged/delicate, and sacred/profane. The scales were divided into seven intervals. Osgood *et al.* gathered data on the ratings of many words by a large number of subjects and then analyzed the data using factor analysis. They discovered that three main factors accounted for most of the variation in the data.

The intuitive meaning of each factor can be understood by looking for the bipolar adjective pairs that are most highly correlated with each factor. The primary factor, which accounted for much of the variation in the data, was highly correlated with good/bad, beautiful/ugly, kind/cruel, and honest/dishonest. Osgood *et al.* called this the *evaluative* factor. The second factor, called the *potency* factor, was highly correlated with strong/weak, large/small, and heavy/light. The third factor, *activity*, was correlated with active/passive, fast/slow, and hot/cold.

In this paper, we focus on the evaluative factor. Hatzivassiloglou and McKeown [1997] call this factor the *semantic orientation* of a word. It is also known as *valence* in the linguistics literature. A positive semantic orientation denotes a positive evaluation (i.e., praise) and a negative semantic orientation denotes a negative evaluation (i.e., criticism). Semantic orientation has both direction (positive or negative) and intensity (mild or strong); contrast okay/fabulous (mild/strong positive) and irksome/horrid (mild/strong negative). We introduce a method for automatically inferring the direction and intensity of the semantic orientation of a word from its statistical association with a set of positive and negative paradigm words.

It is worth noting that there is a high level of agreement among human annotators on the assignment of semantic orientation to words. For their experiments, Hatzivassiloglou and McKeown [1997] created a testing set of 1,336 adjectives (657 positive and 679 negative terms). They labeled the terms themselves and then they validated their labels by asking four people to independently label a random sample of 500 of the 1,336 adjectives. On average, the four people agreed that it was appropriate to assign a positive or negative label to 89% of the 500 adjectives. In the cases where they agreed that it was appropriate to assign a label, they assigned the same label as Hatzivassiloglou and McKeown to 97% of the terms. The average agreement among the four people was also 97%. In our own study, in Section 5.8, the average agreement among the subjects was 98% and the average agreement between the subjects and our benchmark labels was 94% (25 subjects, 28 words). This level of agreement compares favourably with validation studies in similar tasks, such as word sense disambiguation.

This paper presents a general strategy for inferring semantic orientation from semantic association. To provide the motivation for the work described here, Section 2 lists some potential applications of algorithms for determining semantic orientation, such as new kinds of search services [Hearst 1992], filtering "flames" (abusive messages) for newsgroups [Spertus, 1997], and tracking opinions in on-line discussions [Tong, 2001].

Section 3 gives two examples of our method for inferring semantic orientation from association, using two different measures of word association, Pointwise Mutual Information (PMI) [Church and Hanks 1989] and Latent Semantic Analysis (LSA) [Landauer and Dumais 1997]. PMI and LSA are based on co-occurrence, the idea that "a word is characterized by the company it keeps" [Firth 1957]. The hypothesis behind our approach is that the semantic orientation of a word tends to correspond to the semantic orientation of its neighbours.

Related work is examined in Section 4. Hatzivassiloglou and McKeown [1997] have developed a supervised learning algorithm that infers semantic orientation from linguistic constraints on the use of adjectives in conjunctions. The performance of their algorithm was measured by the accuracy with which it classifies words. Another approach is to evaluate an algorithm for learning semantic orientation in the context of a specific application. Turney [2002] does this in the context of text classification, where the task is to classify a review as positive ("thumbs up") or negative ("thumbs down"). Pang *et al.* [2002] have also addressed the task of review classification, but they used standard machine learning text classification techniques.

Experimental results are presented in Section 5. The algorithms are evaluated using 3,596 words (1,614 positive and 1,982 negative) taken from the General Inquirer lexicon [Stone *et al.* 1966]. These words include adjectives, adverbs, nouns, and verbs. An accuracy of 82.8% is attained on the full test set, but the accuracy can rise above 95% when the algorithm is allowed to abstain from classifying mild words.

The interpretation of the experimental results is given in Section 6. We discuss limitations and future work in Section 7 and conclude in Section 8.

## 2. APPLICATIONS

The motivation of Hatzivassiloglou and McKeown [1997] was to use semantic orientation as a component in a larger system, to automatically identify antonyms and distinguish near synonyms. Both synonyms and antonyms typically have strong semantic associations, but synonyms generally have the same semantic orientation, whereas antonyms have opposite orientations.

Semantic orientation may also be used to classify reviews (e.g., movie reviews or automobile reviews) as positive or negative [Turney 2002]. It is possible to classify a review based on the average semantic orientation of phrases in the review that contain adjectives and adverbs. We expect that there will be value in combining semantic orientation [Turney 2002] with more traditional text classification methods for review classification [Pang et al. 2002].

To illustrate review classification, Table 1 shows the average semantic orientation of sentences selected from reviews of banks, from the Epinions site.[1] In this table, we used SO-PMI (see Section 3.1) to calculate the semantic orientation of each individual word and then averaged the semantic orientation of the words in each sentence. Five of these six randomly selected sentences are classified correctly.

Table 1. The average semantic orientation of some sample sentences.

| | Positive Reviews | Average SO |
|---|---|---|
| 1. | I love the local branch, however **communication** may break down if they have to go through head office. | 0.1414 |
| 2. | Bank of America gets my business because of its **extensive** branch and ATM network. | 0.1226 |
| 3. | This bank has exceeded my **expectations** for the last ten years. | 0.1690 |
| | Negative Reviews | Average SO |
| 1. | Do not bank here, their website is even **worse** than their actual locations. | -0.0766 |
| 2. | Use Bank of America only if you like the feeling of a stranger's **warm**, sweaty hands in your pockets. | 0.1535 |
| 3. | If you want **poor** customer service and to lose money to ridiculous charges, Bank of America is for you. | -0.1314 |

In Table 1, for each sentence, the word with the strongest semantic orientation has been marked in bold. These bold words dominate the average and largely determine the orientation of the sentence as a whole. In the sentence that is misclassified as positive, the system is misled by the sarcastic tone. The negative orientations of "stranger's" and "sweaty" were not enough to counter the strong positive orientation of "warm".

---

[1] See http://www.epinions.com/.

One application of review classification is to provide summary statistics for search engines. Given the query "Paris travel review", a search engine could report, "There are 5,000 hits, of which 80% are positive and 20% are negative." The search results could also be sorted by average semantic orientation, so that the user could easily sample the most extreme reviews. Alternatively, the user could include the desired semantic orientation in the query, "Paris travel review orientation: positive" [Hearst 1992].

Preliminary experiments indicate that semantic orientation is also useful for summarization of reviews. A positive review could be summarized by picking out the sentence with the highest positive semantic orientation and a negative review could be summarized by extracting the sentence with the lowest negative semantic orientation.

Another potential application is filtering "flames" for newsgroups [Spertus 1997]. There could be a threshold, such that a newsgroup message is held for verification by the human moderator when the semantic orientation of any word in the message drops below the threshold.

Tong [2001] presents a system for generating *sentiment timelines*. This system tracks online discussions about movies and displays a plot of the number of positive sentiment and negative sentiment messages over time. Messages are classified by looking for specific phrases that indicate the sentiment of the author towards the movie, using a hand-built lexicon of phrases with associated sentiment labels. There are many potential uses for sentiment timelines: Advertisers could track advertising campaigns, politicians could track public opinion, reporters could track public response to current events, and stock traders could track financial opinions. However, with Tong's approach, it would be necessary to provide a new lexicon for each new domain. Tong's [2001] system could benefit from the use of an automated method for determining semantic orientation, instead of (or in addition to) a hand-built lexicon.

Semantic orientation could also be used in an automated chat system (a *chatbot*), to help decide whether a positive or negative response is most appropriate. Similarly, characters in software games would appear more realistic if they responded to the semantic orientation of words that are typed or spoken by the game player.

Another application is the analysis of survey responses to open ended questions. Commercial tools for this task include TextSmart[2] (by SPSS) and Verbatim Blaster[3] (by StatPac). These tools can be used to plot word frequencies or cluster responses into categories, but they do not currently analyze semantic orientation.

---

[2] See http://www.spss.com/textsmart/.
[3] See http://www.statpac.com/content-analysis.htm.

## 3. SEMANTIC ORIENTATION FROM ASSOCIATION

The general strategy in this paper is to infer semantic orientation from semantic association. The semantic orientation of a given word is calculated from the strength of its association with a set of positive words, minus the strength of its association with a set of negative words:

$$Pwords = \text{a set of words with positive semantic orientation} \quad (1)$$

$$Nwords = \text{a set of words with negative semantic orientation} \quad (2)$$

$$A(word_1, word_2) = \text{a measure of association between } word_1 \text{ and } word_2 \quad (3)$$

$$\text{SO-A}(word) = \sum_{pword \in Pwords} A(word, pword) - \sum_{nword \in Nwords} A(word, nword). \quad (4)$$

We assume that $A(word_1, word_2)$ maps to a real number. When $A(word_1, word_2)$ is positive, the words tend to be associated with each other. Larger values correspond to stronger associations. When $A(word_1, word_2)$ is negative, the presence of one word makes it likely that the other is absent.

A word, $word$, is classified as having a positive semantic orientation when SO-A($word$) is positive and a negative orientation when SO-A($word$) is negative. The magnitude (absolute value) of SO-A($word$) can be considered the strength of the semantic orientation.

In the following experiments, seven positive words and seven negative words are used as paradigms of positive and negative semantic orientation:

$$Pwords = \{\text{good, nice, excellent, positive, fortunate, correct, and superior}\} \quad (5)$$

$$Nwords = \{\text{bad, nasty, poor, negative, unfortunate, wrong, and inferior}\}. \quad (6)$$

These fourteen words were chosen for their lack of sensitivity to context. For example, a word such as "excellent" is positive in almost all contexts. The sets also consist of opposing pairs (good/bad, nice/nasty, excellent/poor, etc.). We experiment with randomly selected words in Section 5.8.

It could be argued that this is a supervised learning algorithm with fourteen labeled training examples and millions or billions of unlabeled training examples, but it seems more appropriate to say that the paradigm words are *defining* semantic orientation, rather than *training* the algorithm. Therefore we prefer to describe our approach as unsupervised learning. However, this point does not affect our conclusions.

This general strategy is called SO-A (Semantic Orientation from Association). Selecting particular measures of word association results in particular instances of the strategy. This paper examines SO-PMI (Semantic Orientation from Pointwise Mutual Information) and SO-LSA (Semantic Orientation from Latent Semantic Analysis).

### 3.1. Semantic Orientation from PMI

PMI-IR [Turney 2001] uses Pointwise Mutual Information (PMI) to calculate the strength of the semantic association between words [Church and Hanks 1989]. Word co-occurrence statistics are obtained using Information Retrieval (IR). PMI-IR has been empirically evaluated using 80 synonym test questions from the Test of English as a Foreign Language (TOEFL), obtaining a score of 74% [Turney 2001], comparable to that produced by direct thesaurus search [Littman 2001].

The Pointwise Mutual Information (PMI) between two words, $word_1$ and $word_2$, is defined as follows [Church and Hanks 1989]:

$$\text{PMI}(word_1, word_2) = \log_2\left(\frac{p(word_1 \,\&\, word_2)}{p(word_1)\,p(word_2)}\right). \tag{7}$$

Here, $p(word_1 \,\&\, word_2)$ is the probability that $word_1$ and $word_2$ co-occur. If the words are statistically independent, the probability that they co-occur is given by the product $p(word_1)\,p(word_2)$. The ratio between $p(word_1 \,\&\, word_2)$ and $p(word_1)\,p(word_2)$ is a measure of the degree of statistical dependence between the words. The log of the ratio corresponds to a form of correlation, which is positive when the words tend to co-occur and negative when the presence of one word makes it likely that the other word is absent.

PMI-IR estimates PMI by issuing queries to a search engine (hence the IR in PMI-IR) and noting the number of hits (matching documents). The following experiments use the AltaVista Advanced Search engine[4], which indexes approximately 350 million web pages (counting only those pages that are in English). Given a (conservative) estimate of 300 words per web page, this represents a corpus of at least one hundred billion words.

AltaVista was chosen over other search engines because it has a NEAR operator. The AltaVista NEAR operator constrains the search to documents that contain the words within ten words of one another, in either order. Previous work has shown that NEAR performs better than AND when measuring the strength of semantic association between words [Turney 2001]. We experimentally compare NEAR and AND in Section 5.4.

SO-PMI is an instance of SO-A. From equation (4), we have:

$$\text{SO-PMI}(word) = \sum_{pword \in Pwords} \text{PMI}(word, pword) - \sum_{nword \in Nwords} \text{PMI}(word, nword). \tag{8}$$

---

[4] See http://www.altavista.com/sites/search/adv.

Let hits(*query*) be the number of hits returned by the search engine, given the query, *query*. We calculate PMI(*word₁*, *word₂*) from equation (7) as follows:

$$\text{PMI}(word_1, word_2) = \log_2\left(\frac{\frac{1}{N}\text{hits}(word_1 \text{ NEAR } word_2)}{\frac{1}{N}\text{hits}(word_1) \; \frac{1}{N}\text{hits}(word_2)}\right). \tag{9}$$

Here, *N* is the total number of documents indexed by the search engine. Combining equations (8) and (9), we have:

$$\text{SO-PMI}(word)$$

$$= \log_2\left(\frac{\prod_{pword \in Pwords}\text{hits}(word \text{ NEAR } pword) \cdot \prod_{nword \in Nwords}\text{hits}(nword)}{\prod_{pword \in Pwords}\text{hits}(pword) \cdot \prod_{nword \in Nwords}\text{hits}(word \text{ NEAR } nword)}\right). \tag{10}$$

Note that *N*, the total number of documents, drops out of the final equation. Equation (10) is a log-odds ratio [Agresti 1996].

Calculating the semantic orientation of a word via equation (10) requires twenty-eight queries to AltaVista (assuming there are fourteen paradigm words). Since the two products in (10) that do not contain *word* are constant for all words, they only need to be calculated once. Ignoring these two constant products, the experiments required only fourteen queries per word.

To avoid division by zero, 0.01 was added to the number of hits. This is a form of Laplace smoothing. We examine the effect of varying this parameter in Section 5.3.

Pointwise Mutual Information is only one of many possible measures of word association. Several others are surveyed in Manning and Schütze [1999]. Dunning [1993] suggests the use of likelihood ratios as an improvement over PMI. To calculate likelihood ratios for the association of two words, *X* and *Y*, we need to know four numbers:

$k(X\ Y)$ = the frequency that *X* occurs within a given neighbourhood of *Y* (11)

$k(\sim X\ Y)$ = the frequency that *Y* occurs in a neighbourhood without *X* (12)

$k(X \sim Y)$ = the frequency that *X* occurs in a neighbourhood without *Y* (13)

$k(\sim X \sim Y)$ = the frequency that neither *X* nor *Y* occur in a neighbourhood. (14)

If the neighbourhood size is ten words, then we can use hits(*X* NEAR *Y*) to estimate $k(X\ Y)$ and hits(*X*) – hits(*X* NEAR *Y*) to estimate $k(X \sim Y)$, but note that these are only rough estimates, since hits(*X* NEAR *Y*) is the number of *documents* that contain *X* near *Y*, not the number of *neighbourhoods* that contain *X* and *Y*. Some preliminary experiments suggest that this distinction is important, since alternatives to PMI (such as likelihood ratios [Dunning 1993] and the Z-score [Smadja 1993]) appear to perform worse than PMI when used with search engine hit counts.

However, if we do not restrict our attention to measures of word association that are compatible with search engine hit counts, there are many possibilities. In the next subsection, we look at one of them, Latent Semantic Analysis.

## 3.2. Semantic Orientation from LSA

SO-LSA applies Latent Semantic Analysis (LSA) to calculate the strength of the semantic association between words [Landauer and Dumais 1997]. LSA uses the Singular Value Decomposition (SVD) to analyze the statistical relationships among words in a corpus.

The first step is to use the text to construct a matrix $\mathbf{X}$, in which the row vectors represent words and the column vectors represent chunks of text (e.g., sentences, paragraphs, documents). Each cell represents the *weight* of the corresponding word in the corresponding chunk of text. The weight is typically the tf-idf score (Term Frequency times Inverse Document Frequency) for the word in the chunk. (tf-idf is a standard tool in information retrieval [van Rijsbergen 1979].)[5]

The next step is to apply singular value decomposition [Golub and Van Loan 1996] to $\mathbf{X}$, to decompose $\mathbf{X}$ into a product of three matrices $\mathbf{U\Sigma V}^T$, where $\mathbf{U}$ and $\mathbf{V}$ are in column orthonormal form (i.e., the columns are orthogonal and have unit length: $\mathbf{U}^T\mathbf{U} = \mathbf{V}^T \mathbf{V} = \mathbf{I}$) and $\mathbf{\Sigma}$ is a diagonal matrix of *singular values* (hence SVD). If $\mathbf{X}$ is of rank $r$, then $\mathbf{\Sigma}$ is also of rank $r$. Let $\mathbf{\Sigma}_k$, where $k < r$, be the diagonal matrix formed from the top $k$ singular values, and let $\mathbf{U}_k$ and $\mathbf{V}_k$ be the matrices produced by selecting the corresponding columns from $\mathbf{U}$ and $\mathbf{V}$. The matrix $\mathbf{U}_k\mathbf{\Sigma}_k\mathbf{V}_k^T$ is the matrix of rank $k$ that best approximates the original matrix $\mathbf{X}$, in the sense that it minimizes the approximation errors. That is, $\hat{\mathbf{X}} = \mathbf{U}_k\mathbf{\Sigma}_k\mathbf{V}_k^T$ minimizes $\|\hat{\mathbf{X}} - \mathbf{X}\|_F$ over all matrices $\hat{\mathbf{X}}$ of rank $k$, where $\|\cdots\|_F$ denotes the Frobenius norm [Golub and Van Loan 1996; Bartell *et al.* 1992]. We may think of this matrix $\mathbf{U}_k\mathbf{\Sigma}_k\mathbf{V}_k^T$ as a "smoothed" or "compressed" version of the original matrix $\mathbf{X}$.

LSA is similar to principal components analysis. LSA works by measuring the similarity of words using the smoothed matrix $\mathbf{U}_k\mathbf{\Sigma}_k\mathbf{V}_k^T$ instead of the original matrix $\mathbf{X}$. The similarity of two words, LSA($word_1$, $word_2$), is measured by the cosine of the angle between their corresponding row vectors in $\mathbf{U}_k\mathbf{\Sigma}_k\mathbf{V}_k^T$, which is equivalent to using the corresponding rows of $\mathbf{U}_k$ [Deerwester *et al.* 1990; Bartell *et al.* 1992; Schütze 1993; Landauer and Dumais 1997].

The semantic orientation of a word, *word*, is calculated by SO-LSA from equation (4), as follows:

$$\text{SO-LSA}(word) = \sum_{pword \in Pwords} \text{LSA}(word, pword) - \sum_{nword \in Nwords} \text{LSA}(word, nword). \quad (15)$$

For the paradigm words, we have the following (from equations (5), (6), and (15)):

$$\begin{aligned}\text{SO-LSA}(word) &= [\text{LSA}(word, \text{good}) + ... + \text{LSA}(word, \text{superior})] \\ &\quad - [\text{LSA}(word, \text{bad}) + ... + \text{LSA}(word, \text{inferior})].\end{aligned} \quad (16)$$

As with SO-PMI, a word, *word*, is classified as having a positive semantic orientation when SO-LSA(*word*) is positive and a negative orientation when SO-LSA(*word*) is negative. The magnitude of SO-LSA(*word*) represents the strength of the semantic orientation.

## 4. RELATED WORK

Related work falls into three groups: work on classifying words by positive or negative semantic orientation (Section 4.1), classifying reviews (e.g., movie reviews) as positive or negative (Section 4.2), and recognizing subjectivity in text (Section 4.3).

### 4.1. Classifying Words

Hatzivassiloglou and McKeown [1997] treat the problem of determining semantic orientation as a problem of classifying words, as we also do in this paper. They note that there are linguistic constraints on the semantic orientations of adjectives in conjunctions. As an example, they present the following three sentences:

1. The tax proposal was simple and well received by the public.
2. The tax proposal was simplistic, but well received by the public.
3. (*) The tax proposal was simplistic and well received by the public.

The third sentence is incorrect, because we use "and" with adjectives that have the same semantic orientation ("simple" and "well-received" are both positive), but we use "but" with adjectives that have different semantic orientations ("simplistic" is negative).

Hatzivassiloglou and McKeown [1997] use a four-step supervised learning algorithm to infer the semantic orientation of adjectives from constraints on conjunctions:

1. All conjunctions of adjectives are extracted from the given corpus.

---

[5] The tf-idf score gives more weight to terms that are statistically "surprising". This heuristic works well for information retrieval, but its impact on determining semantic orientation is unknown.

2. A supervised learning algorithm combines multiple sources of evidence to label pairs of adjectives as having the same semantic orientation or different semantic orientations. The result is a graph where the nodes are adjectives and links indicate sameness or difference of semantic orientation.
3. A clustering algorithm processes the graph structure to produce two subsets of adjectives, such that links across the two subsets are mainly different-orientation links, and links inside a subset are mainly same-orientation links.
4. Since it is known that positive adjectives tend to be used more frequently than negative adjectives, the cluster with the higher average frequency is classified as having positive semantic orientation.

For brevity, we will call this the HM algorithm.

Like SO-PMI and SO-LSA, HM can produce a real-valued number that indicates both the direction (positive or negative) and the strength of the semantic orientation. The clustering algorithm (Step 3 above) can produce a "goodness-of-fit" measure that indicates how well an adjective fits in its assigned cluster.

Hatzivassiloglou and McKeown [1997] used a corpus of 21 million words and evaluated HM with 1,336 manually labeled adjectives (657 positive and 679 negative). Their results are given in Table 2. HM classifies adjectives with accuracies ranging from 78% to 92%, depending on Alpha, as described next.

Table 2. The accuracy of HM with a 21 million-word corpus.[6]

| Alpha | Accuracy | Size of test set | Percent of "full" test set |
|---|---|---|---|
| 2 | 78.08% | 730 | 100.0% |
| 3 | 82.56% | 516 | 70.7% |
| 4 | 87.26% | 369 | 50.5% |
| 5 | 92.37% | 236 | 32.3% |

Alpha is a parameter that is used to partition the 1,336 labeled adjectives into training and testing sets. As Alpha increases, the training set grows and the testing set becomes smaller. The precise definition of Alpha is complicated, but the basic idea is to put the hard cases (the adjectives for which there are few conjunctions in the given corpus) in the training set and the easy cases (the adjectives for which there are many conjunctions) in the testing set. As Alpha increases, the testing set becomes increasingly easy (that is, the adjectives that remain in the testing set are increasingly well covered by the given corpus). In essence, the idea is to improve accuracy by abstaining from classifying the difficult (rare, sparsely represented) adjectives. As expected, the accuracy rises as Alpha rises. This suggests that the accuracy will improve with larger corpora.

---

[6] This table is derived from Table 3 in Hatzivassiloglou and McKeown [1997].

This algorithm is able to achieve good accuracy levels, but it has some limitations. In contrast with SO-A, HM is restricted to adjectives and it requires labeled adjectives as training data (in step 2).

Although each step in HM, taken by itself, is relatively simple, the combination of the four steps makes theoretical analysis challenging. In particular, the interaction between the supervised labeling (step 2) and the clustering (step 3) is difficult to analyze. For example, the degree of regularization (i.e., smoothing, pruning) in the labeling step may have an impact on the quality of the clusters. By contrast, SO-PMI is captured in a single formula (equation (10)), which takes the form of the familiar log-odds ratio [Agresti 1996].

HM has only been evaluated with adjectives, but it seems likely that it would work with adverbs. For example, we would tend to say "He ran quickly (+) *but* awkwardly (–)" rather than "He ran quickly (+) *and* awkwardly (–)". However, it seems less likely that HM would work well with nouns and verbs. There is nothing wrong with saying "the rise (+) *and* fall (–) of the Roman Empire" or "love (+) *and* death (–)".[7] Indeed, "but" would not work in these phrases.

Kamps and Marx [2002] use the WordNet lexical database [Miller 1990] to determine the semantic orientation of a word. For a given word, they look at its semantic distance from "good" compared to its semantic distance from "bad". The idea is similar to SO-A, except that the measure of association is replaced with a measure of semantic distance, based on WordNet [Budanitsky and Hirst 2001]. This is an interesting approach, but it has not yet been evaluated empirically.

### 4.2. Classifying Reviews

Turney [2002] used a three-step approach to classify reviews. The first step was to apply a part-of-speech tagger to the review and then extract two-word phrases, such as "romantic ambience" or "horrific events", where one of the words in the phrase was an adjective or an adverb. The second step was to use SO-PMI to calculate the semantic orientation of each extracted phrase. The third step was to classify the review as positive or negative, based on the average semantic orientation of the extracted phrases. If the average was positive, then the review was classified as positive; otherwise, negative. The experimental results suggest that SO-PMI may be useful for classifying reviews, but the results do not reveal how well SO-PMI can classify individual words or phrases. Therefore it is worthwhile to experimentally evaluate the performance of SO-PMI on individual words, as we do in Section 5.

---

[7] *The Rise and Fall of the Roman Empire* is the title of a book by Edward Gibbon. *Love and Death* is the title of a movie directed by Woody Allen.

The reviewing application of SO-A illustrates the value of an automated approach to determining semantic orientation. Although it might be feasible to manually create a lexicon of individual words labeled with semantic orientation, if an application requires the semantic orientation of two-word or three-word phrases, the number of terms involved grows beyond what can be handled by manual labeling. Turney [2002] observed that an adjective such as "unpredictable" may have a negative semantic orientation in an automobile review, in a phrase such as "unpredictable steering", but it could have a positive (or neutral) orientation in a movie review, in a phrase such as "unpredictable plot". SO-PMI can handle multiword phrases by simply searching for them using a quoted phrase query.

Pang *et al.* [2002] applied classical text classification techniques to the task of classifying movie reviews as positive or negative. They evaluated three different supervised learning algorithms and eight different sets of features, yielding twenty-four different combinations. The best result was achieved using a Support Vector Machine (SVM) with features based on the presence or absence (rather than the frequency) of single words (rather than two-word phrases).

We expect that Pang *et al.*'s algorithm will tend to be more accurate than Turney's, since the former is supervised and the latter is unsupervised. On the other hand, we hypothesize that the supervised approach will require retraining for each new domain. For example, if a supervised algorithm is trained with movie reviews, it is likely to perform poorly when it is tested with automobile reviews. Perhaps it is possible to design a hybrid algorithm that achieves high accuracy without requiring retraining.

Classifying reviews is related to measuring semantic orientation, since it is one of the possible applications for semantic orientation, but there are many other possible applications (see Section 2). Although it is interesting to evaluate a method for inferring semantic orientation, such as SO-PMI, in the context of an application, such as review classification, the diversity of potential applications makes it interesting to study semantic orientation in isolation, outside of any particular application. That is the approach adopted in this paper.

### 4.3. Subjectivity Analysis

Other related work is concerned with determining subjectivity [Hatzivassiloglou and Wiebe 2000; Wiebe 2000; Wiebe *et al.* 2001]. The task is to distinguish sentences (or paragraphs or documents or other suitable chunks of text) that present opinions and evaluations from sentences that objectively present factual information [Wiebe 2000].

Wiebe *et al.* [2001] list a variety of potential applications for automated subjectivity tagging, such as recognizing "flames" [Spertus, 1997], classifying email, recognizing speaker role in radio broadcasts, and mining reviews. In several of these applications, the first step is to recognize that the text is subjective and then the natural second step is to determine the semantic orientation of the subjective text. For example, a flame detector cannot merely detect that a newsgroup message is subjective, it must further detect that the message has a negative semantic orientation; otherwise a message of praise could be classified as a flame.

On the other hand, applications that involve semantic orientation are also likely to benefit from a prior step of subjectivity analysis. For example, a movie review typically contains a mixture of objective descriptions of scenes in the movie and subjective statements of the viewer's reaction to the movie. In a positive movie review, it is common for the objective description to include words with a negative semantic orientation, although the subjective reaction may be quite positive [Turney 2002]. If the task is to classify the review as positive or negative, a two-step approach seems wise. The first step would be to filter out the objective sentences [Wiebe 2000; Wiebe *et al.* 2001] and the second step would be to determine the semantic orientation of the words and phrases in the remaining subjective sentences [Turney 2002].

### 5. EXPERIMENTS

In Section 5.1, we discuss the lexicons and corpora that are used in the following experiments. Section 5.2 examines the baseline performance of SO-PMI, when it is configured as described in Section 3.1. Sections 5.3, 5.4, and 5.5 explore variations on the baseline SO-PMI system. The baseline performance of SO-LSA is evaluated in Section 5.6 and variations on the baseline SO-LSA system are considered in Section 5.7. The final experiments in Section 5.8 analyze the effect of the choice of the paradigm words, for both SO-PMI and SO-LSA.

## 5.1. Lexicons and Corpora

The following experiments use two different lexicons and three different corpora. The corpora are used for unsupervised learning and the lexicons are used to evaluate the results of the learning. The *HM lexicon* is the list of 1,336 labeled adjectives that was created by Hatzivassiloglou and McKeown [1997]. The *GI lexicon* is a list of 3,596 labeled words extracted from the General Inquirer lexicon [Stone *et al.* 1966]. The *AV-ENG corpus* is the set of English web pages indexed by the AltaVista search engine. The *AV-CA corpus* is the set of English web pages in the Canadian domain that are indexed by AltaVista. The *TASA corpus* is a set of short English documents gathered from a variety of sources by Touchstone Applied Science Associates.

The HM lexicon consists of 1,336 adjectives, 657 positive and 679 negative [Hatzivassiloglou and McKeown 1997]. We described this lexicon earlier, in Sections 1 and 4.1. We use the HM lexicon to allow comparison between the approach of Hatzivassiloglou and McKeown [1997] and the SO-A algorithms described here.

Since the HM lexicon is limited to adjectives, most of the following experiments use a second lexicon, the GI lexicon, which consists of 3,596 adjectives, adverbs, nouns, and verbs, 1,614 positive and 1,982 negative [Stone *et al.* 1966]. The General Inquirer lexicon is available at http://www.wjh.harvard.edu/~inquirer/. The lexicon was developed by Philip Stone and his colleagues, beginning in the 1960's, and continues to grow. It has been designed as a tool for *content analysis*, a technique used by social scientists, political scientists, and psychologists for objectively identifying specified characteristics of messages [Stone *et al.* 1966].

The full General Inquirer lexicon has 182 categories of word tags and 11,788 words. The words tagged "Positiv" (1,915 words) and "Negativ" (2,291 words) have (respectively) positive and negative semantic orientations. Table 3 lists some examples.

Table 3. Examples of "Positiv" and "Negativ" words.

| Positiv | | Negativ | |
|---|---|---|---|
| abide | absolve | abandon | abhor |
| ability | absorbent | abandonment | abject |
| able | absorption | abate | abnormal |
| abound | abundance | abdicate | abolish |

Words with multiple senses may have multiple entries in the lexicon. The list of 3,596 words (1,614 positive and 1,982 negative) used in the subsequent experiments was generated by reducing multiple-entry words to single entries. Some words with multiple senses were tagged as both "Positiv" and "Negativ". For example, "mind" in the sense of "intellect" is positive, but "mind" in the sense of "beware" is negative. These ambiguous words were not included in our set of 3,596 words. We also excluded the fourteen paradigm words (good/bad, nice/nasty, etc.).

Of the words in the HM lexicon, 47.7% also appear in the GI lexicon (324 positive, 313 negative). The agreement between the two lexicons on the orientation of these shared words is 98.3% (6 terms are positive in HM but negative in GI; 5 terms are negative in HM but positive in GI).

The AltaVista search engine is available at http://www.altavista.com/. Based on estimates in the popular press and our own tests with various queries, we estimate that the AltaVista index contained approximately 350 million English web pages at the time our experiments were carried out. This corresponds to roughly one hundred billion words. We call this the AV-ENG corpus. The set of web pages indexed by AltaVista is constantly changing, but there is enough stability that our experiments were reliably repeatable over the course of several months.

In order to examine the effect of corpus size on learning, we used AV-CA, a subset of the AV-ENG corpus. The AV-CA corpus was produced by adding "AND host:.ca" to every query to AltaVista, which restricts the search results to the web pages with "ca" in the host domain name. This consists mainly of hosts that end in "ca" (the Canadian domain), but it also includes a few hosts with "ca" in other parts of the domain name (such as "http://www.ca.com/"). The AV-CA corpus contains approximately 7 million web pages (roughly two billion words), about 2% of the size of the AV-ENG corpus.

Our experiments with SO-LSA are based on the online demonstration of LSA, available at http://lsa.colorado.edu/. This demonstration allows a choice of several different corpora. We chose the largest corpus, the TASA-ALL corpus, which we call simply TASA. In the online LSA demonstration, TASA is called the "General Reading up to 1st year college (300 factors)" topic space. The corpus contains a wide variety of short documents, taken from novels, newspaper articles, and other sources. It was collected by Touchstone Applied Science Associates, to develop The Educator's Word Frequency Guide. The TASA corpus contains approximately 10 million words, about 0.5% of the size of the AV-CA corpus.

The TASA corpus is not indexed by AltaVista. For SO-PMI, the following experimental results were generated by emulating AltaVista on a local copy of the TASA corpus. We used a simple Perl script to calculate the hits() function for TASA, as a surrogate for sending queries to AltaVista.

5.2. SO-PMI Baseline

Table 4 shows the accuracy of SO-PMI in its baseline configuration, as described in Section 3.1. These results are for all three corpora, tested with the HM lexicon. In this table, the strength (absolute value) of the semantic orientation was used as a measure of confidence that the word will be correctly classified. Test words were sorted in descending order of the absolute value of their semantic orientation and the top ranked words (the highest confidence words) were then classified. For example, the second row in Table 4 shows the accuracy when the top 75% (with highest confidence) were classified and the last 25% (with lowest confidence) were ignored.

Table 4. The accuracy of SO-PMI with the HM lexicon and the three corpora.

| Percent of full test set | Size of test set | Accuracy with AV-ENG | Accuracy with AV-CA | Accuracy with TASA |
|---|---|---|---|---|
| 100% | 1336 | 87.13% | 80.31% | 61.83% |
| 75% | 1002 | 94.41% | 85.93% | 64.17% |
| 50% | 668 | 97.60% | 91.32% | 46.56% |
| 25% | 334 | 98.20% | 92.81% | 70.96% |
| Approx. num. of words in corpus | | $1 \times 10^{11}$ | $2 \times 10^9$ | $1 \times 10^7$ |

The performance of SO-PMI in Table 4 can be compared to the performance of the HM algorithm in Table 2 (Section 4.1), since both use the HM lexicon, but there are some differences in the evaluation, since the HM algorithm is supervised but SO-PMI is unsupervised. Because the HM algorithm is supervised, part of the HM lexicon must be set aside for training, so the algorithm cannot be evaluated on the whole lexicon. Aside from this caveat, it appears that the performance of the HM algorithm is roughly comparable to the performance of SO-PMI with the AV-CA corpus, which is about one hundred times larger than the corpus used by Hatzivassiloglou and McKeown [1997] ($2 \times 10^9$ words versus $2 \times 10^7$ words). This suggests that the HM algorithm makes more efficient use of corpora than SO-PMI, but the advantage of SO-PMI is that it can easily be scaled up to very large corpora, where it can achieve significantly higher accuracy.

The results of these experiments are shown in more detail in Figure 1. The percentage of the full test set (labeled *threshold* in the figure) varies from 5% to 100% in increments of 5%. Three curves are plotted, one for each of the three corpora. The figure shows that a smaller corpus not only results in lower accuracy, but also results in less stability. With the larger corpora, the curves are relatively smooth; with the smallest corpus, the curve looks quite noisy.

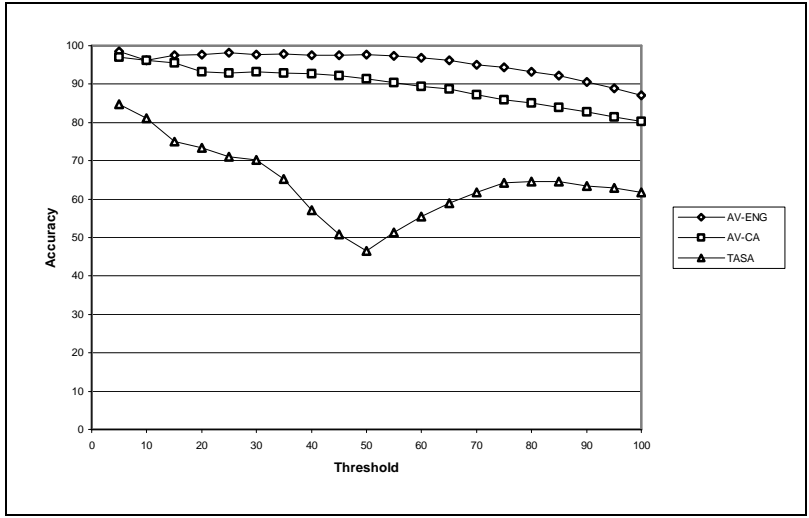

Figure 1. Accuracy of SO-PMI with the HM lexicon and the three corpora.

Table 5 shows the accuracy of SO-PMI with the GI lexicon, which includes adverbs, nouns, and verbs, in addition to adjectives. Figure 2 gives more detail. Compared with Table 4 and Figure 1, there is a slight drop in accuracy, but the general trends are the same.

Table 5. The accuracy of SO-PMI with the GI lexicon and the three corpora.

| Percent of full test set | Size of test set | Accuracy with AV-ENG | Accuracy with AV-CA | Accuracy with TASA |
|---|---|---|---|---|
| 100% | 3596 | 82.84% | 76.06% | 61.26% |
| 75% | 2697 | 90.66% | 81.76% | 63.92% |
| 50% | 1798 | 95.49% | 87.26% | 47.33% |
| 25% | 899 | 97.11% | 89.88% | 68.74% |
| Approx. num. of words in corpus | | $1 \times 10^{11}$ | $2 \times 10^9$ | $1 \times 10^7$ |

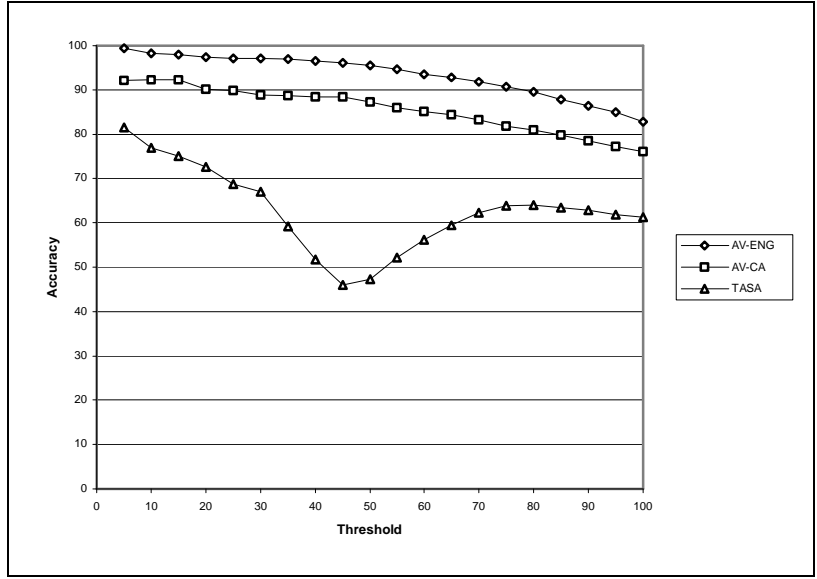

Figure 2. Accuracy of SO-PMI with the GI lexicon and the three corpora.

### 5.3. Varying the Laplace Smoothing Factor

As we mentioned in Section 3.1, we used a Laplace smoothing factor of 0.01 in the baseline version of SO-PMI. In this section, we explore the impact of varying the smoothing factor.

Figure 3 graphs the accuracy of SO-PMI as a function of the smoothing factor, which varies from 0.0001 to 10,000 (note the logarithmic scale), using the AV-ENG corpus and the GI lexicon. There are four curves, for four different thresholds on the percentage of the full test set that is classified. The smoothing factor has relatively little impact until it rises above 10, at which point the accuracy begins to fall off. The optimal value is about 1, although the difference between 1 and 0.1 or 0.01 is slight.

Figure 4 shows the same experimental setup, except using the AV-CA corpus. We see the same general pattern, but the accuracy begins to decline a little earlier, when the smoothing factor rises above 0.1. The highest accuracy is attained when the smoothing factor is about 0.1. The AV-CA corpus (approximately $2 \times 10^9$ words) is more sensitive to the smoothing factor than the AV-ENG corpus (approximately $1 \times 10^{11}$ words). A smoothing factor of about 0.1 seems to help SO-PMI handle the increased noise, due to the smaller corpus (compare Figure 3 and Figure 4).

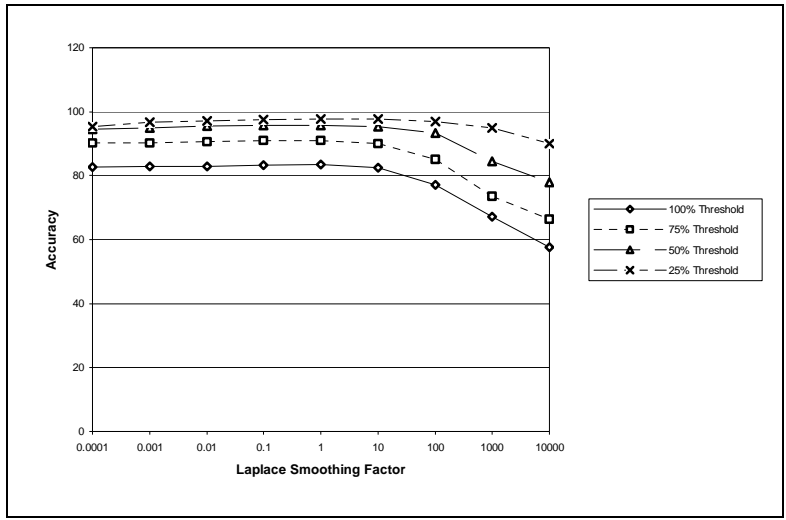

Figure 3. Effect of Laplace smoothing factor with AV-ENG and the GI lexicon.

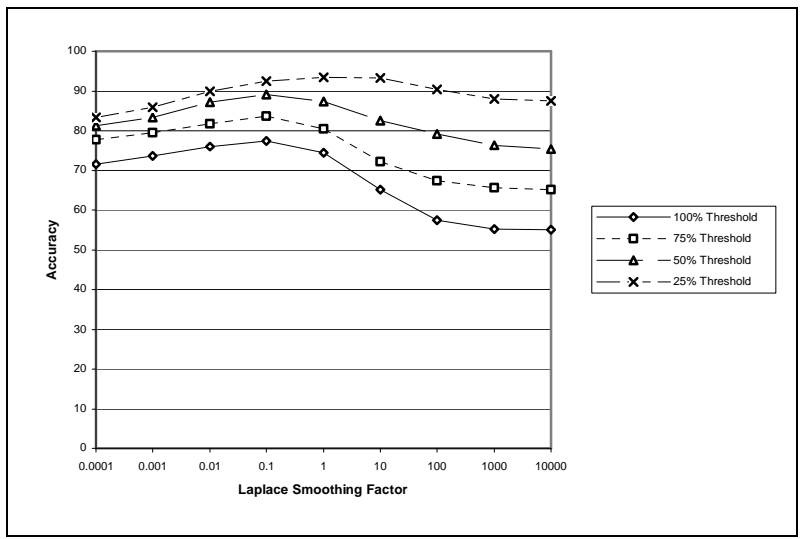

Figure 4. Effect of Laplace smoothing factor with AV-CA and the GI lexicon.

Figure 5 plots the performance with varying smoothing factors using the smallest corpus, TASA. The performance is quite sensitive to the choice of smoothing factor. Our baseline value of 0.01 turns out to be a poor choice for the TASA corpus. The optimal value is about 0.001. This suggests that, when using SO-PMI with a small corpus, it would be wise to use cross-validation to optimize the value of the Laplace smoothing factor.

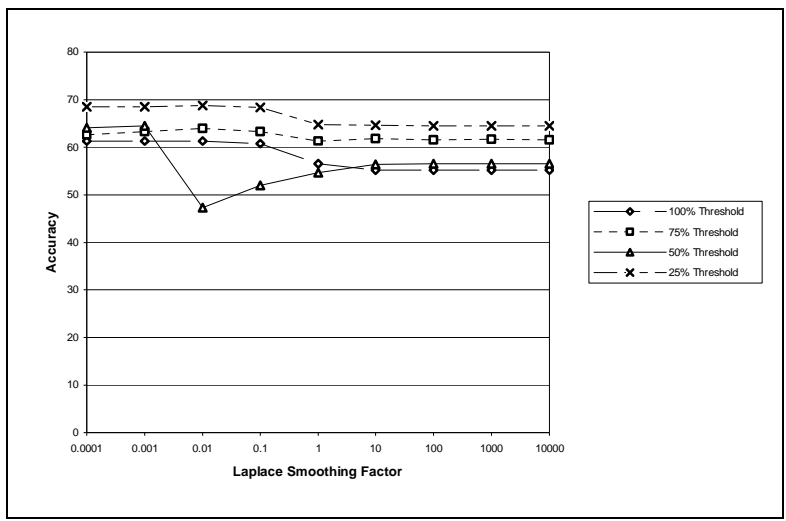

Figure 5. Effect of Laplace smoothing factor with TASA and the GI lexicon.

These three figures show that the optimal smoothing factor increases as the size of the corpus increases, as expected. The figures also show that the impact of the smoothing factor decreases as the corpus size increases. There is less need for smoothing when a large quantity of data is available. The baseline smoothing factor of 0.01 was chosen to avoid division by zero, not to provide resistance to noise. The benefit from optimizing the smoothing factor for noise resistance is small for large corpora.

### 5.4. Varying the Neighbourhood Size

The AltaVista NEAR operator restricts search to a fixed neighbourhood of ten words, but we can vary the neighbourhood size with the TASA corpus, since we have a local copy of the corpus. Figure 6 shows accuracy as a function of the neighbourhood size, as we vary the size from 2 to 1000 words, using TASA and the GI lexicon.

The advantage of a small neighbourhood is that words that occur closer to each other are more likely to be semantically related. The disadvantage is that, for any pair of words, there will usually be more occurrences of the pair within a large neighbourhood than within a small neighbourhood, so a larger neighbourhood will tend to have higher statistical reliability. An optimal neighbourhood size will balance these conflicting effects. A larger corpus should yield better statistical reliability than a smaller corpus, so the optimal neighbourhood size will be smaller with a larger corpus. The optimal neighbourhood size will also be determined by the frequency of the words in the test set. Rare words will favour a larger neighbourhood size than frequent words.

Figure 6 shows that, for the TASA corpus and the GI lexicon, it seems best to have a neighbourhood size of at least 100 words. The TASA corpus is relatively small, so it is not surprising that a large neighbourhood size is best. The baseline neighbourhood size of 10 words is clearly suboptimal for TASA.

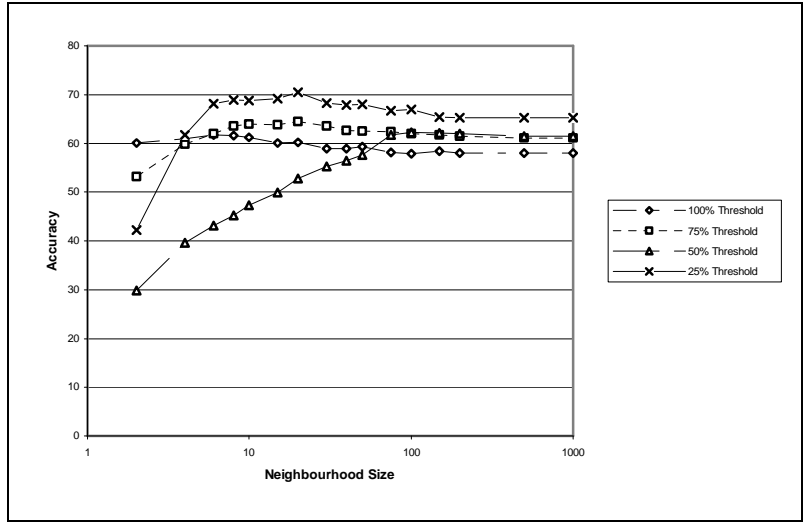

Figure 6. SO-PMI with varying neighbourhoods, using TASA and the GI lexicon.

With AltaVista, we can use the AND operator instead of the NEAR operator, which increases the neighbourhood size from ten words to a whole document. Figure 7 is a graph of accuracy as a function of the percentage of the test set that is classified (threshold), using AV-ENG and the GI lexicon. With the whole test set, NEAR is clearly superior to AND, but the gap closes as the threshold decreases. This is not surprising, since, as the threshold decreases, the selected testing words have increasingly high confidences. That is, the absolute values of the semantic orientations of the remaining words grow increasingly large. The words with a very strong semantic orientation (high absolute value) do not need the extra sensitivity of NEAR; they are easily classified using the less sensitive AND operator.

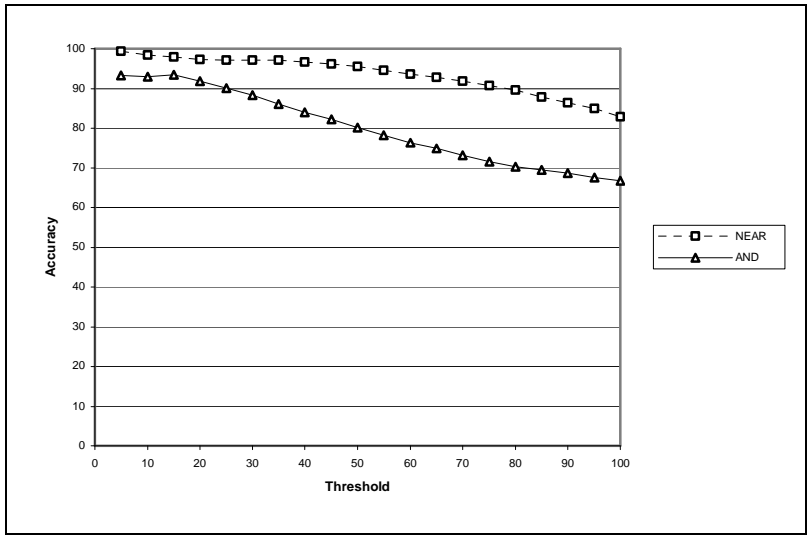

Figure 7. AND versus NEAR with AV-ENG and the GI lexicon.

Figure 8, with AV-CA ($2 \times 10^9$ words), displays the same general pattern as Figure 7, with AV-ENG ($1 \times 10^{11}$ words). However, on the smaller corpus, AND is superior to NEAR for the words with the strongest semantic orientation (threshold below 10%). The smaller corpus shows more clearly the trade-off between the greater sensitivity of a small neighbourhood (with NEAR) and the greater resistance to noise of a large neighbourhood (with AND).

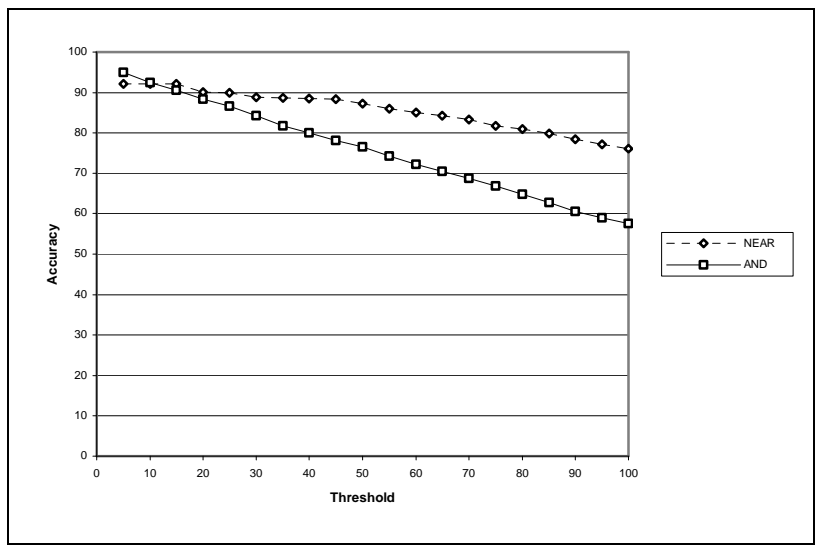

Figure 8. AND versus NEAR with AV-CA and the GI lexicon.

## 5.5. Product versus Disjunction

Recall equation (10), for calculating SO-PMI(*word*):

$$\text{SO-PMI}(word) = \log_2 \left( \frac{\prod_{pword \in Pwords} \text{hits}(word \text{ NEAR } pword) \cdot \prod_{nword \in Nwords} \text{hits}(nword)}{\prod_{pword \in Pwords} \text{hits}(pword) \cdot \prod_{nword \in Nwords} \text{hits}(word \text{ NEAR } nword)} \right). \quad (17)$$

As we discussed in Section 3.1, this equation requires fourteen queries to AltaVista for each word (ignoring the constant terms). In this section, we investigate whether the number of queries can be reduced by combining the paradigm words, using the OR operator.

For convenience, we introduce the following definitions:

$$Pquery = \underset{pword \in Pwords}{\text{OR}} pword \quad (18)$$

$$Nquery = \underset{nword \in Nwords}{\text{OR}} nword. \quad (19)$$

Given the fourteen paradigm words, for example, we have the following (from equations (5), (6), (18), and (19)):

$$Pquery = (\text{good OR nice OR ... OR superior}) \quad (20)$$

$$Nquery = (\text{bad OR nasty OR ... OR inferior}). \quad (21)$$

We attempt to approximate (17) as follows:[8]

$$\text{SO-PMI}(word) = \log_2 \left( \frac{\text{hits}(word \text{ NEAR } Pquery) \cdot \text{hits}(Nquery)}{\text{hits}(word \text{ NEAR } Nquery) \cdot \text{hits}(Pquery)} \right). \quad (22)$$

Calculating the semantic orientation of a word using equation (22) requires only two queries per word, instead of fourteen (ignoring the constant terms, hits(*Pquery*) and hits(*Nquery*)).

Figure 9 plots the performance of product (equation (17)) versus disjunction (equation (22)) for SO-PMI with the AV-ENG corpus and the GI lexicon. Figure 10 shows the performance with the AV-CA corpus and Figure 11 with the TASA corpus. For the largest corpus, there is a clear advantage to using our original equation (17), but the two equations have similar performance with the smaller corpora. Since the execution time of SO-PMI is almost completely dependent on the number of queries sent to AltaVista, equation (22) executes seven times faster than equation (17). Therefore the disjunction equation should be preferred for smaller corpora and the product equation should be preferred for larger corpora.

---
[8] We use OR here, because using AND or NEAR would almost always result in zero hits. We add 0.01 to the hits, to avoid division by zero.

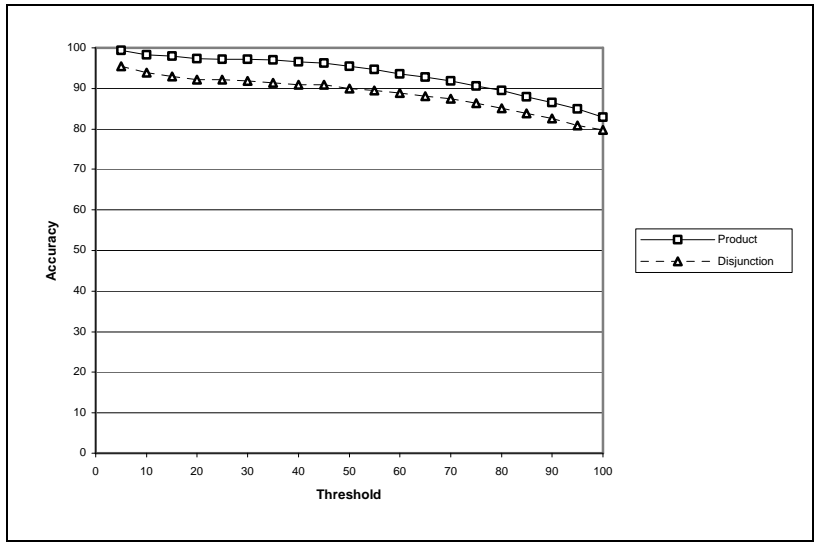

Figure 9. Accuracy of product versus disjunction with AV-ENG and GI.

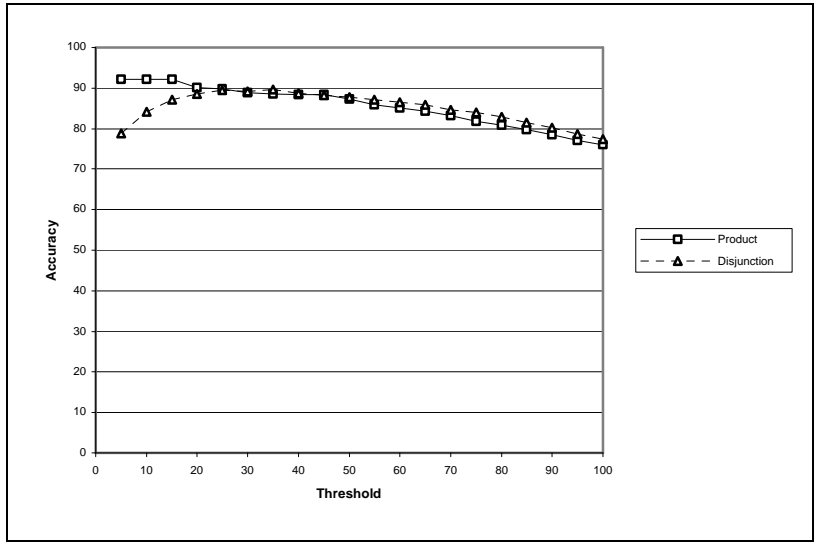

Figure 10. Accuracy of product versus disjunction with AV-CA and GI.

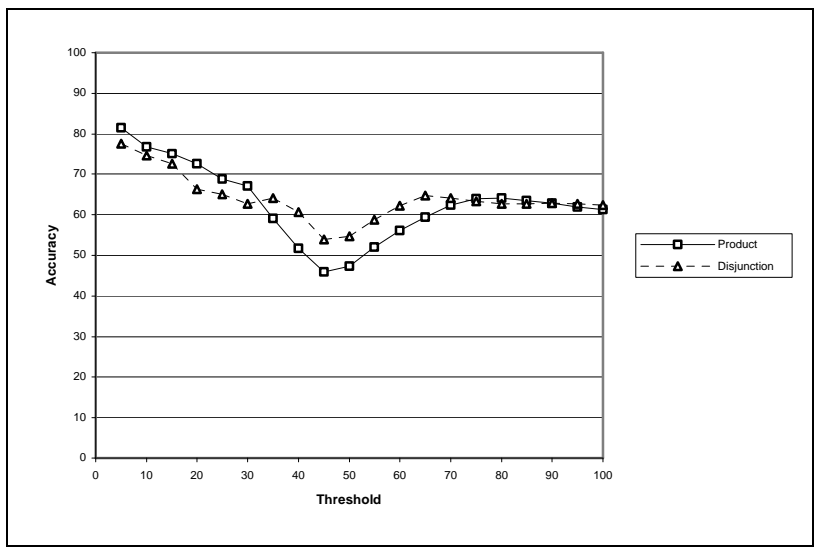

Figure 11. Accuracy of product versus disjunction with TASA and GI.

## 5.6. SO-LSA Baseline

Table 6 shows the performance of SO-LSA on TASA with the HM lexicon. The experiment used the online demonstration of LSA, mentioned in Section 5.1. The TASA corpus was used to generate a matrix **X** with 92,409 rows (words) and 37,651 columns (each document in TASA corresponds to one column), and SVD was used to reduce the matrix to 300 dimensions. This is the baseline configuration of SO-LSA, as described in Section 3.2.

Table 6. The accuracy of SO-LSA and SO-PMI with the HM lexicon and TASA.

| Percent of full test set | Size of test set | Accuracy of SO-LSA | Accuracy of SO-PMI |
|---|---|---|---|
| 100% | 1336 | 67.66% | 61.83% |
| 75% | 1002 | 73.65% | 64.17% |
| 50% | 668 | 79.34% | 46.56% |
| 25% | 334 | 88.92% | 70.96% |

For ease of comparison, Table 6 also gives the performance of SO-PMI on TASA with the HM lexicon, copied from Table 4. LSA has not yet been scaled up to corpora of the sizes of AV-ENG or AV-CA, so we cannot compare SO-LSA and SO-PMI on these larger corpora. Figure 12 presents a more detailed comparison, as the threshold varies from 5% to 100% in increments of 5%.

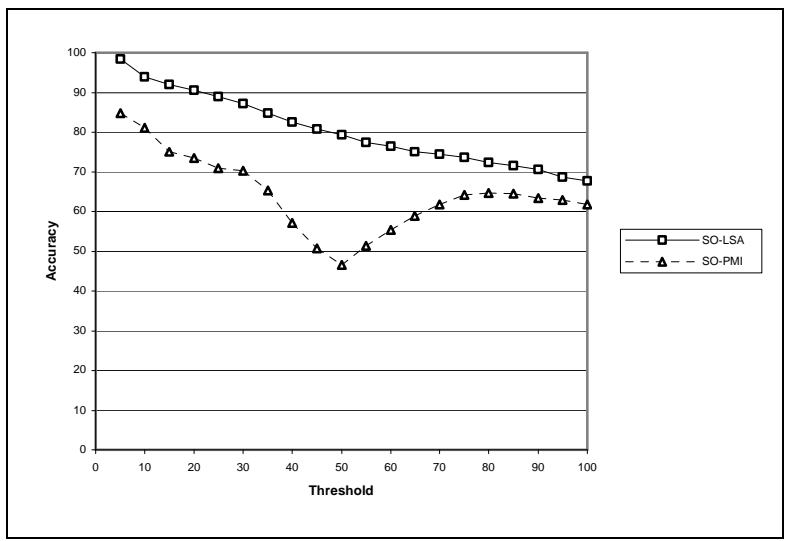

Figure 12. Comparison of SO-LSA and SO-PMI with the HM lexicon and TASA.

Table 7 and Figure 13 give the corresponding results for the GI lexicon. The accuracy is slightly lower with the GI lexicon, but we see the same general trend as with the HM lexicon. SO-PMI and SO-LSA have approximately the same accuracy when evaluated on the full test set (threshold 100%), but SO-LSA rapidly pulls ahead as we decrease the percentage of the test set that is classified. It appears that the magnitude of SO is a better indicator of confidence for SO-LSA than for SO-PMI, at least when the corpus is relatively small.

Table 7. The accuracy of SO-LSA and SO-PMI with the GI lexicon and TASA.

| Percent of full test set | Size of test set | Accuracy of SO-LSA | Accuracy of SO-PMI |
| --- | --- | --- | --- |
| 100% | 3596 | 65.27% | 61.26% |
| 75% | 2697 | 71.04% | 63.92% |
| 50% | 1798 | 75.58% | 47.33% |
| 25% | 899 | 81.98% | 68.74% |

In addition to its lower accuracy, SO-PMI appears less stable than SO-LSA, especially as the threshold drops below 75%. Comparing with Figure 6, we see that, although a larger neighbourhood makes SO-PMI more stable, even a neighbourhood of 1000 words (which is like using AND with AltaVista) will not bring SO-PMI up the accuracy levels of SO-LSA.

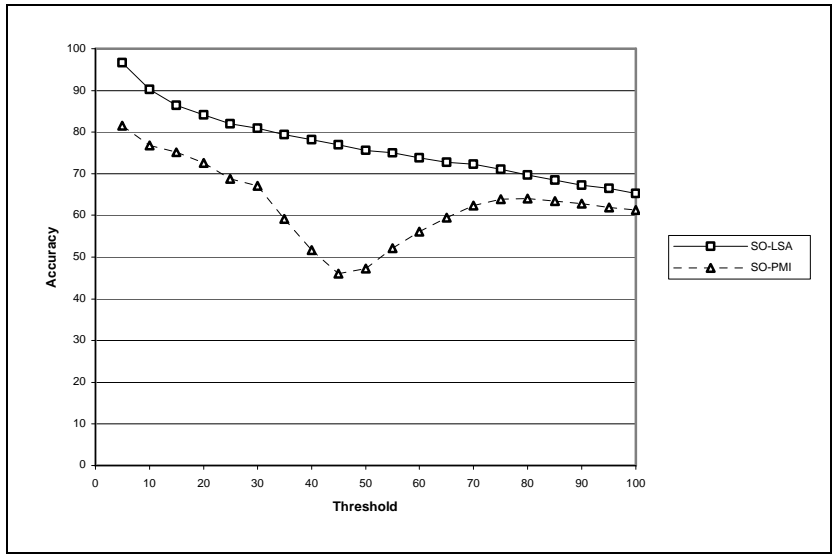

Figure 13. Comparison of SO-LSA and SO-PMI with the GI lexicon and TASA.

## 5.7. Varying the Number of Dimensions

The behaviour of LSA is known to be sensitive to the number of dimensions of the matrix (the parameter $k$ in Section 3.2). In this section, we investigate the effect of varying the number of dimensions for SO-LSA with the TASA corpus and the GI lexicon. Figure 14 shows the accuracy of SO-LSA as a function of the number of dimensions. The $k$ parameter varies from 50 to 300 dimensions, in increments of 50. The highest accuracy is achieved with 250 dimensions. Second highest is 200 dimensions, followed by 300 dimensions. The graph suggests that the optimal value of $k$, for using SO-LSA with the TASA corpus, is somewhere between 200 and 300 dimensions, likely near 250 dimensions.

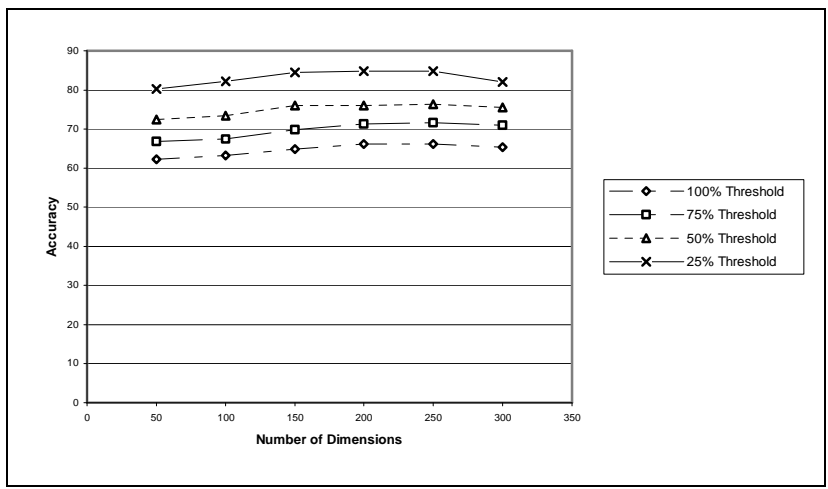

Figure 14. The effect of varying the number of dimensions for SO-LSA.

## 5.8. Varying the Paradigm Words

The standard methodology for supervised learning is to randomly split the labeled data (the lexicon, in this context) into a training set and a testing set. The sizes of the training and testing sets are usually approximately the same, within an order of magnitude. We think of SO-A as an unsupervised learning method, because the "training" set is only fourteen words (two orders of magnitude smaller than the testing set) and because the paradigm words were carefully chosen instead of randomly selected (*defining* rather than *training*).

The fourteen paradigm words were chosen as prototypes or ideal examples of positive and negative semantic orientation (see Section 3). All fourteen paradigm words appear in the General Inquirer lexicon. The positive paradigm words are all tagged "Positiv" and the negative paradigm words are all tagged "Negativ" (although they were chosen before consulting the General Inquirer lexicon). As we mentioned, the paradigm words were removed from the testing words for our experiments.

The following experiment examines the behaviour of SO-A when the paradigm words are randomly selected. Since rare words would tend to require a larger corpus for SO-A to work well, we controlled for frequency effects. For each original paradigm word, we found the word in the General Inquirer lexicon with the same tag ("Positiv" or "Negativ") and the most similar frequency. The frequency was measured by the number of hits in AltaVista. Table 8 shows the resulting new paradigm words.

Table 8. Original paradigm words and corresponding frequency-matched new paradigm words.

| Original paradigm word | Frequency of original word | Matched new word | Frequency of new word | Semantic orientation |
|---|---|---|---|---|
| good | 55,289,359 | right | 55,321,211 | positive |
| nice | 12,259,779 | worth | 12,242,455 | positive |
| excellent | 11,119,032 | commission | 11,124,607 | positive |
| positive | 9,963,557 | classic | 9,969,619 | positive |
| fortunate | 1,049,242 | devote | 1,052,922 | positive |
| correct | 11,316,975 | super | 11,321,807 | positive |
| superior | 5,335,487 | confidence | 5,344,805 | positive |
| bad | 18,577,687 | lost | 17,962,401 | negative |
| nasty | 2,273,977 | burden | 2,267,307 | negative |
| poor | 9,622,080 | pick | 9,660,275 | negative |
| negative | 5,896,695 | raise | 5,885,800 | negative |
| unfortunate | 987,942 | guilt | 989,363 | negative |
| wrong | 12,048,581 | capital | 11,721,649 | negative |
| inferior | 1,013,356 | blur | 1,011,693 | negative |

The inclusion of some of the words in Table 8, such as "pick", "raise", and "capital", may seem surprising. These words are only negative in certain contexts, such as "pick on your brother", "raise a protest", and "capital offense". We hypothesized that the poor performance of the new paradigm words was (at least partly) due to their sensitivity to context, in contrast to the original paradigm words. To test this hypothesis, we asked 25 people to rate the 28 words in Table 8, using the following scale:

1 = negative semantic orientation (in almost all contexts)
2 = negative semantic orientation (in typical contexts)
3 = neutral or context-dependent semantic orientation
4 = positive semantic orientation (in typical contexts)
5 = positive semantic orientation (in almost all contexts)

Each person was given a different random permutation of the 28 words, to control for ordering effects. The average pairwise correlation between subjects' ratings was 0.86. The original paradigm words had average ratings of 4.5 for the seven positive words and 1.4 for the seven negative words. The new paradigm words had average ratings of 3.9 for positive and 2.4 for negative. These judgments lend support to the hypothesis that context sensitivity is higher for the new paradigm words; context independence is higher for the original paradigm words. On an individual basis, subjects judged the original word more context independent than the corresponding new paradigm word in 61% of cases (statistically significant, $p < .01$).

To evaluate the fourteen new paradigm words, we removed them from the set of 3,596 testing words and substituted the original paradigm words in their place. Figure 15 compares the accuracy of the original paradigm words with the new words, using SO-PMI with AV-ENG and GI, and Figure 16 uses AV-CA. It is clear that the original words perform much better than the new words.

Figure 17 and Figure 18 compare SO-PMI and SO-LSA on the TASA-ALL corpus with the original and new paradigm words. Again, the original words perform much better than the new words.

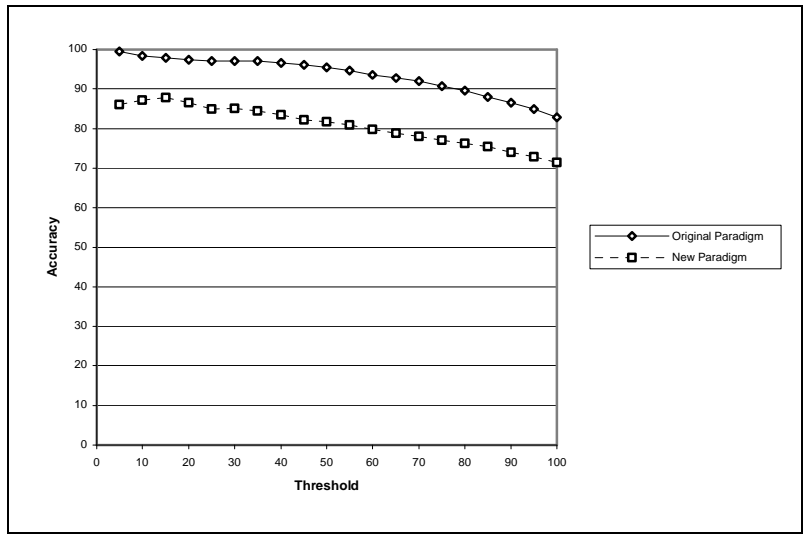

Figure 15. Original paradigm versus new, using SO-PMI with AV-ENG and GI.

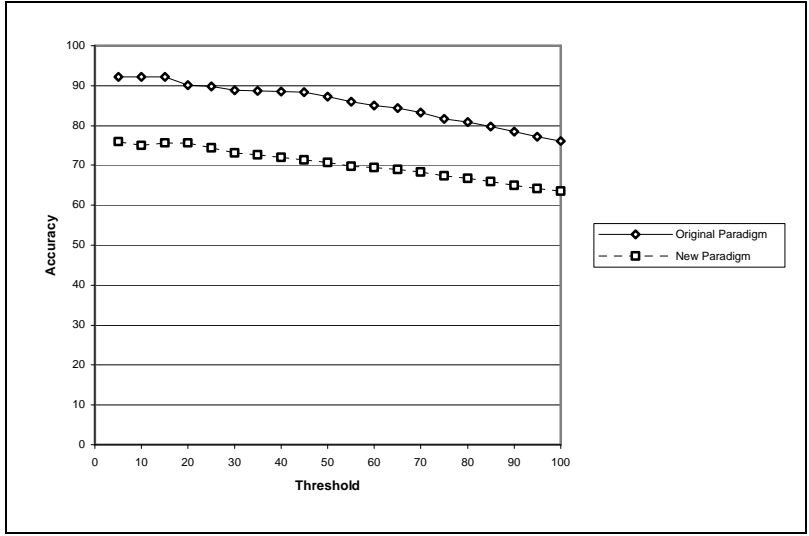

Figure 16. Original paradigm versus new, using SO-PMI with AV-CA and GI.

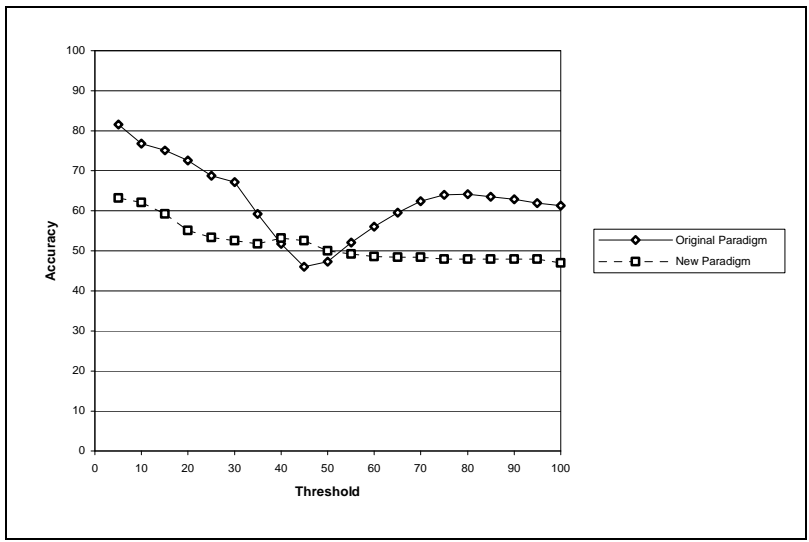

Figure 17. Original paradigm versus new, using SO-PMI with TASA and GI.

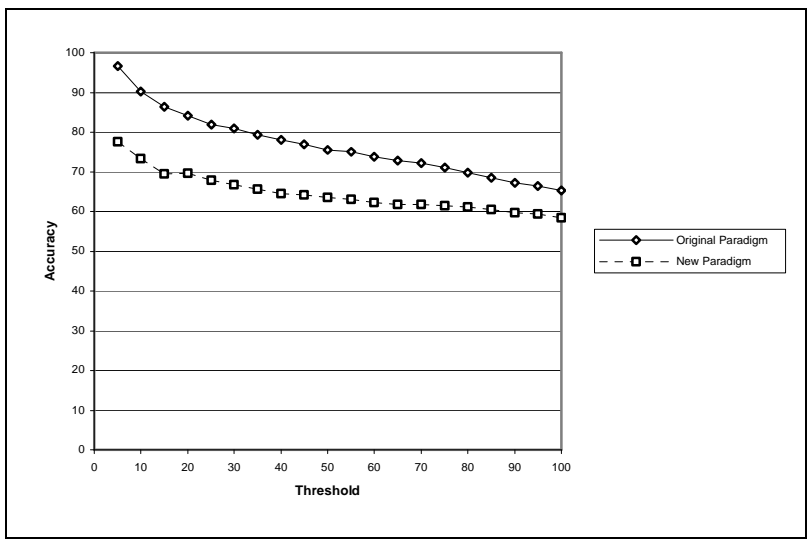

Figure 18. Original paradigm versus new, using SO-LSA with TASA and GI.

## 6. DISCUSSION OF RESULTS

LSA has not yet been scaled up to corpora of the sizes that are available for PMI-IR, so we were unable to evaluate SO-LSA on the larger corpora that were used to evaluate SO-PMI. However, the experiments suggest that SO-LSA is able to use data more efficiently than SO-PMI, and SO-LSA might surpass the accuracy attained by SO-PMI with AV-ENG, given a corpus of comparable size.

PMI measures the degree of association between two words by the frequency with which they co-occur. That is, if PMI($word_1$, $word_2$) is positive, then $word_1$ and $word_2$ tend to occur near each other. Resnik [1995] argues that such *word-word* co-occurrence approaches are able to capture "relatedness" of words, but do not specifically address similarity of meaning. LSA, on the other hand, measures the degree of association between two words by comparing the contexts in which the two words occur. That is, if LSA($word_1$, $word_2$) is positive, then (in general) there are many words, $word_i$, such that $word_1$ tends to occur near $word_i$ and $word_2$ tends to occur near $word_i$. It appears that such *word-context* co-occurrence approaches correlate better with human judgments of semantic similarity than word-word co-occurrence approaches [Landauer 2002]. This could help explain LSA's apparent efficiency of data usage.

Laplace smoothing was used in SO-PMI primarily to prevent division by zero, rather than to provide resistance to noise, which is why the relatively small value of 0.01 was chosen. The experiments show that the performance of SO-PMI is not particularly sensitive to the value of the smoothing factor with larger corpora.

The size of the neighbourhood for SO-PMI seems to be an important parameter, especially when the corpus is small. For the TASA corpus, a neighbourhood size of 1000 words (which is the same as a whole document, since the largest document is 650 words long) yields the best results. On the other hand, for the larger corpora, a neighbourhood size of ten words (NEAR) results in higher accuracy than using the whole document (AND). For best results, it seems that the neighbourhood size should be tuned for the given corpus and the given test words (rarer test words will tend to need larger neighbourhoods).

Given the TASA corpus and the GI lexicon, SO-LSA appears to work best with a 250 dimensional space. This is approximately the same number as other researchers have found useful in other applications of LSA [Deerwester *et al.* 1990; Landauer and Dumais 1997]. However, the accuracy with 200 or 300 dimensions is almost the same as the accuracy with 250 dimensions; SO-LSA is not especially sensitive to the value of this parameter.

The experiments with alternative paradigm words show that both SO-PMI and SO-LSA are sensitive to the choice of paradigm words. It appears that the difference between the original paradigm words and the new paradigm words is that the former are less context-sensitive. Since SO-A estimates semantic orientation by association with the paradigm words, it is not surprising that it is important to use paradigm words that are robust, in the sense that their semantic orientation is relatively insensitive to context.

## 7. LIMITATIONS AND FUTURE WORK

A limitation of SO-A is the size of the corpora required for good performance. A large corpus of text requires significant disk space and processing time. In our experiments with SO-PMI, we paused for five seconds between each query, as a courtesy to AltaVista. Processing the 3,596 words taken from the General Inquirer lexicon required 50,344 queries, which took about 70 hours. This can be reduced to 10 hours, using equation (22) instead of equation (17), but there may be a loss of accuracy, as we saw in Section 5.5.

However, improvements in hardware will reduce the impact of this limitation. In the future, corpora of a hundred billion words will be common and the average desktop computer will be able to process them easily. Today, we can indirectly work with corpora of this size through web search engines, as we have done in this paper. With a little bit of creativity, a web search engine can tell us a lot about language use.

The ideas in SO-A can likely be extended to many other semantic aspects of words. The General Inquirer lexicon has 182 categories of word tags [Stone *et al.* 1966] and this paper has only used two of them, so there is no shortage of future work. For example, another interesting pair of categories in General Inquirer is *strong* and *weak*. Although *strong* tends to be correlated with *positive* and *weak* with *negative*, there are many examples in General Inquirer of words that are *negative* and *strong* (e.g., abominable, aggressive, antagonism, attack, austere, avenge) or *positive* and *weak* (e.g., delicate, gentle, modest, polite, subtle). The *strong/weak* pair may be useful in applications such as analysis of political text, propaganda, advertising, news, and opinions. Many of the applications discussed in Section 2 could also make use of the ability to automatically distinguish *strong* and *weak* words.

As we discussed in Section 5.8, the semantic orientation of many words depends on the context. For example, in the General Inquirer lexicon, mind#9 ("lose one's mind") is Negativ and mind#10 ("right mind") is Positiv. In our experiments, we avoided this issue by deleting words like "mind", with both Positiv and Negativ tags, from the set of testing words. However, in a real-world application, the issue cannot be avoided so easily.

This may appear to be a problem of word sense disambiguation. Perhaps, in one sense, the word "mind" is positive and, in another sense, it is negative. Although it is related to word sense disambiguation, we believe that it is a separate problem. For example, consider "unpredictable steering" versus "unpredictable plot" (from Section 4.2). The word "unpredictable" has the same meaning in both phrases, yet it has a negative orientation in the first case but a positive orientation in the second case. We believe that the problem is context sensitivity. This is supported by the experiments in Section 5.8. Evaluating the semantic orientation of two-word phrases, instead of single words, is an attempt to deal with this problem [Turney 2002], but more sophisticated solutions might yield significant improvements in performance, especially with applications that involve larger chunks of text (e.g., paragraphs and documents instead of words and phrases).

## 8. CONCLUSION

This paper has presented a general strategy for measuring semantic orientation from semantic association, SO-A. Two instances of this strategy have been empirically evaluated, SO-PMI and SO-LSA. SO-PMI requires a large corpus, but it is simple, easy to implement, unsupervised, and it is not restricted to adjectives.

Semantic orientation has a wide variety of applications in information systems, including classifying reviews, distinguishing synonyms and antonyms, extending the capabilities of search engines, summarizing reviews, tracking opinions in online discussions, creating more responsive chatbots, and analyzing survey responses. There are likely to be many other applications that we have not anticipated.

ACKNOWLEDGEMENTS

Thanks to the anonymous reviewers of ACM TOIS for their very helpful comments. We are grateful to Vasileios Hatzivassiloglou and Kathy McKeown for generously providing a copy of their lexicon. Thanks to Touchstone Applied Science Associates for the TASA corpus. We thank AltaVista for allowing us to send so many queries to their search engine. Thanks to Philip Stone and his colleagues for making the General Inquirer lexicon available to researchers. We would also like to acknowledge the support of NASA and Knowledge Engineering Technologies.

## 9. REFERENCES

AGRESTI, A. 1996. *An introduction to categorical data analysis*. Wiley, New York.

BARTELL, B.T., COTTRELL, G.W., AND BELEW, R.K. 1992. Latent semantic indexing is an optimal special case of multidimensional scaling. *Proceedings of the Fifteenth Annual International ACM SIGIR Conference on Research and Development in Information Retrieval*, 161-167.

BUDANITSKY, A. AND HIRST, G. 2001. Semantic distance in WordNet: An experimental, application-oriented evaluation of five measures. *Workshop on WordNet and Other Lexical Resources, Second meeting of the North American Chapter of the Association for Computational Linguistics*, Pittsburgh, PA.

CHURCH, K.W., AND HANKS, P. 1989. Word association norms, mutual information and lexicography. *Proceedings of the 27th Annual Conference of the Association of Computational Linguistics*. Association for Computational Linguistics, New Brunswick, NJ, 76-83.

DEERWESTER, S., DUMAIS, S.T., FURNAS, G.W., LANDAUER, T.K., AND HARSHMAN, R. 1990. Indexing by latent semantic analysis. *Journal of the American Society for Information Science*, 41(6), 391-407.

DUNNING, T. 1993. Accurate methods for the statistics of surprise and coincidence. *Computational Linguistics*, 19, 61-74.

FIRTH, J.R. 1957. *A Synopsis of Linguistic Theory 1930-1955*. In *Studies in Linguistic Analysis*, Philological Society, Oxford, 1-32. Reprinted in F.R. Palmer (ed.), *Selected Papers of J.R. Firth 1952-1959*, Longman, London, 1968.

GOLUB, G.H., AND VAN LOAN, C.F. 1996. *Matrix Computations*. Third edition. Johns Hopkins University Press, Baltimore, MD.

HATZIVASSILOGLOU, V., AND MCKEOWN, K.R. 1997. Predicting the semantic orientation of adjectives. *Proceedings of the 35th Annual Meeting of the Association for Computational Linguistics and the 8th Conference of the European Chapter of the ACL*. Association for Computational Linguistics, New Brunswick, NJ, 174-181.

HATZIVASSILOGLOU, V., AND WIEBE, J.M. 2000. Effects of adjective orientation and gradability on sentence subjectivity. *Proceedings of 18th International Conference on Computational Linguistics*. Association for Computational Linguistics, New Brunswick, NJ.

HEARST, M.A. 1992. Direction-based text interpretation as an information access refinement. In P. Jacobs (Ed.), *Text-Based Intelligent Systems: Current Research and Practice in Information Extraction and Retrieval*. Lawrence Erlbaum Associates, Mahwah, NJ.

KAMPS, J., AND MARX, M. 2002. Words with attitude. *Proceedings of the First International Conference on Global WordNet*, CIIL, Mysore, India, 332-341.

LANDAUER, T.K., AND DUMAIS, S.T. 1997. A solution to Plato's problem: The latent semantic analysis theory of the acquisition, induction, and representation of knowledge. *Psychological Review,* 104, 211-240.

LANDAUER, T.K. 2002. On the computational basis of learning and cognition: Arguments from LSA. To appear in B.H. Ross (Ed.), *The Psychology of Learning and Motivation*.

LITTMAN, M.L. 2001. Language games and other meaningful pursuits. Presentation slides. (http://www.cs.rutgers.edu/~mlittman/talks/CA-lang.ppt).

MANNING, C.D., AND SCHÜTZE, H. 1999. *Foundations of Statistical Natural Language Processing*. MIT Press, Cambridge, MA.

MILLER, G.A. 1990. WordNet: An on-line lexical database. *International Journal of Lexicography*, 3(4), 235-312.

OSGOOD, C.E., SUCI, G.J., AND TANNENBAUM, P.H. 1957. *The Measurement of Meaning*. University of Illinois Press, Chicago.

PANG, B., LEE, L., AND VAITHYANATHAN, S. 2002. Thumbs up? Sentiment classification using machine learning techniques. *Proceedings of the 2002 Conference on Empirical Methods in Natural Language Processing*, 79-86.

RESNIK, P. 1995. Using information content to evaluate semantic similarity in a taxonomy. *Proceedings of the 14th International Joint Conference on Artificial Intelligence*. Morgan Kaufmann, San Mateo, CA, 448-453.

SCHÜTZE, H. 1993. Word space. In S. J. Hanson, J. D. Cowan, and C. L. Giles, editors, *Advances in Neural Information Processing Systems 5*. Morgan Kaufmann., San Mateo, CA, 895-902.

SMADJA, F. 1993. Retrieving collocations from Text: Xtract. *Computational Linguistics*, 19, 143-177.

SPERTUS, E. 1997. Smokey: Automatic recognition of hostile messages. *Proceedings of the Conference on Innovative Applications of Artificial Intelligence*. AAAI Press, Menlo Park, CA, 1058-1065.

STONE, P. J., DUNPHY, D. C., SMITH, M. S., AND OGILVIE, D. M. 1966. *The General Inquirer: A Computer Approach to Content Analysis.* MIT Press, Cambridge, MA.

TONG, R.M. 2001. An operational system for detecting and tracking opinions in on-line discussions. *Working Notes of the ACM SIGIR 2001 Workshop on Operational Text Classification*. ACM, New York, NY, 1-6.

TURNEY, P.D. 2001. Mining the Web for synonyms: PMI-IR versus LSA on TOEFL. *Proceedings of the Twelfth European Conference on Machine Learning*. Springer-Verlag, Berlin, 491-502.

TURNEY, P.D. 2002. Thumbs up or thumbs down? Semantic orientation applied to unsupervised classification of reviews. *Proceedings of the Association for Computational Linguistics 40th Anniversary Meeting*. Association for Computational Linguistics, New Brunswick, NJ.

VAN RIJSBERGEN, C.J. 1979. *Information Retrieval* (2nd edition), Butterworths, London.

WIEBE, J.M. 2000. Learning subjective adjectives from corpora. *Proceedings of the 17th National Conference on Artificial Intelligence*. AAAI Press, Menlo Park, CA.

WIEBE, J.M., BRUCE, R., BELL, M., MARTIN, M., & WILSON, T. 2001. A corpus study of evaluative and speculative language. *Proceedings of the Second ACL SIG on Dialogue Workshop on Discourse and Dialogue*. Aalborg, Denmark.